\documentclass[letterpaper, 10 pt, journal, twoside]{ieeetran}
\usepackage[utf8]{inputenc}
\IEEEoverridecommandlockouts    

\usepackage{multicol}
\usepackage[bookmarks=true]{hyperref} 

\usepackage[detect-weight=true, detect-family=true]{siunitx}

\newtheorem{definition}{Definition}
\newtheorem{assumption}{Assumption}

\usepackage{xcolor}
\usepackage{amsmath}
\usepackage{bbm}
\usepackage{graphicx}
\usepackage{amssymb}
\usepackage[detect-weight=true, detect-family=true]{siunitx}
\usepackage[normalem]{ulem}

\DeclareSIUnit{\million}{M}
\DeclareSIUnit{\thousand}{k}

\usepackage{float}
\usepackage{censor}
\usepackage{nicefrac}

\usepackage[ruled,vlined,linesnumbered]{algorithm2e}
\SetInd{0.25em}{0.5em}
\SetKwRepeat{Do}{do}{while}

\usepackage{dsfont}

\usepackage{tikz}
\usetikzlibrary{fit,calc}
\newcommand*{\tikzmk}[1]{\tikz[remember picture,overlay,] \node (#1) {};\ignorespaces}

\newcommand{\markline}[1]{\tikz[remember picture,overlay]{\node[yshift=3pt,xshift=#1,fill=yellow!100,opacity=.25,fit={(A)($(A)+(0.95\linewidth,-2.5\baselineskip)$)},rounded corners=4pt] {};}\ignorespaces}

\usepackage[natbib=true,
    style=ieee,
    citestyle=numeric-comp,
    backend=biber,
    maxbibnames=5,
    maxcitenames=99,
    doi=false,
    url=false,
    giveninits=true]{biblatex}
\DeclareSourcemap{
  \maps[datatype=bibtex]{
    \map{
      \step[fieldsource=url,
        match=\regexp{\\_},
        replace=\regexp{_}]
    }
  }
}

\AtBeginBibliography{\footnotesize}
\newbibmacro{string+url}[1]{%
  \iffieldundef{url}{%
  \iffieldundef{doi}{#1}{\href{http://dx.doi.org/\thefield{doi}}{#1}}}{\href{\thefield{url}}{#1}}}
\DeclareFieldFormat*{title}{\usebibmacro{string+url}{``#1''}}

\makeatletter
\let\MYcaption\@makecaption
\makeatother

\usepackage[font=footnotesize]{subcaption}

\makeatletter
\let\@makecaption\MYcaption
\makeatother

\DeclareMathOperator*{\argmax}{arg\,max}
\DeclareMathOperator*{\argmin}{arg\,min}

\def\HiLi{\leavevmode\rlap{\hbox to \hsize{\color{yellow!50}\leaders\hrule height .8\baselineskip depth .5ex\hfill}}}

\newcommand{\vs}{\mathbf{s}}
\newcommand{\va}{\mathbf{a}}
\newcommand{\vz}{\mathbf{z}}
\newcommand{\vg}{\mathbf{g}}

\newcommand{\vp}{\mathbf{p}}
\newcommand{\vv}{\mathbf{v}}
\newcommand{\vy}{\mathbf{y}}
\newcommand{\vmu}{\mathbf{\mu}}
\newcommand{\vSigma}{\mathbf{\Sigma}}

\newcommand{\cH}{H}
\newcommand{\cN}{N}

\newcommand{\cdelta}{\Delta}
\newcommand{\cphysicalradius}{r_p}
\newcommand{\csensingradius}{r_\mathrm{sense}}
\newcommand{\cgoalradius}{r_g}
\newcommand{\ctagradius}{r_t}
\newcommand{\cpul}{\overline{p}} 

\newcommand{\cvul}{\overline{v}}

\newcommand{\caul}{\overline{a}}

\newcommand{\gG}{\mathcal{G}}

\newcommand{\sI}{\mathcal{I}} 
 
\newcommand{\sS}{\mathcal{S}}
\newcommand{\sA}{\mathcal{A}}
\newcommand{\sZ}{\mathcal{Z}}
\newcommand{\sX}{\mathcal{X}}
\newcommand{\sU}{\mathcal{U}}
\newcommand{\sD}{\mathcal{D}}
\newcommand{\sN}{\mathcal{N}}

\newcommand{\fT}{\mathcal{T}}
\newcommand{\fR}{\mathcal{R}}
\newcommand{\fO}{\mathcal{O}}
\newcommand{\fV}{\mathit{V}}
\newcommand{\ff}{\mathit{f}}
\newcommand{\fh}{\mathit{h}}
\newcommand{\fpi}{\mathit{\pi}}
\newcommand{\fL}{\mathcal{L}}

\newcommand{\fhv}{\tilde{V}} 
\newcommand{\fhpi}{\tilde{\pi}} 
\newcommand{\fepi}{\pi^e} 
\newcommand{\flpi}{\pi^{le}} 
\newcommand{\fpioptimal}{\pi^*} 

\renewcommand{\th}{^{\text{th}}}

\newcommand{\onode}{\mathrm{n}}

\begin{document}

\title{
Neural Tree Expansion for Multi-Robot Planning in Non-Cooperative Environments
}


\author{Benjamin Rivi\`ere$^{1}$, Wolfgang  H\"onig$^{1}$, Matthew Anderson$^{1}$, and Soon-Jo Chung$^{1}$, 
\thanks{Manuscript received: February, 21, 2021; Revised May 26, 2021; Accepted June 24, 2021.}
\thanks{This paper was recommended for publication by Editor Ani Hsieh upon evaluation of the Associate Editor and Reviewers' comments.
This work was supported by Defense Advanced Research Projects Agency (DARPA). The views, opinions and/or findings expressed are those of the authors and should not be interpreted as representing the official views or policies of the Department of Defense or the U.S. Government. Preliminary work was in part funded by Raytheon. Video: \url{https://youtu.be/mklbTfWl7DE}. Code: \url{https://github.com/bpriviere/decision_making}.} 
\thanks{$^{1}$ Graduate Aerospace Laboratories of the California Institute of Technology
        {\tt\footnotesize \texttt{\{briviere, whoenig, matta, sjchung\}@caltech.edu}.}}%
\thanks{Digital Object Identifier (DOI): see top of this page.}
}

\maketitle

\begin{abstract}
We present a self-improving, Neural Tree Expansion (NTE) method for multi-robot online planning in non-cooperative environments, where each robot attempts to maximize its cumulative reward while interacting with other self-interested robots.
Our algorithm adapts the centralized, perfect information, discrete-action space method from AlphaZero to a decentralized, partial information, continuous action space setting for multi-robot applications. 
Our method has three interacting components: (i) a centralized, perfect-information ``expert'' Monte Carlo Tree Search (MCTS) with large computation resources that provides expert demonstrations, (ii) a decentralized, partial-information ``learner'' MCTS with small computation resources that runs in real-time and provides self-play examples, and (iii) policy \& value neural networks that are trained with the expert demonstrations and bias both the expert and the learner tree growth. 
Our numerical experiments demonstrate Neural Tree Expansion's computational advantage by finding better solutions than a MCTS with \si{20} times more resources.
The resulting policies are dynamically sophisticated, demonstrate coordination between robots, and play the Reach-Target-Avoid differential game significantly better than the state-of-the-art control-theoretic baseline for multi-robot, double-integrator systems. Our hardware experiments on an aerial swarm demonstrate the computational advantage of Neural Tree Expansion, enabling online planning at \SI{20}{Hz} with effective policies in complex scenarios. 
\end{abstract}


\IEEEpeerreviewmaketitle

\section{Introduction}

\IEEEPARstart{M}{ulti-agent} interactions in non-cooperative environments are ubiquitous in robotic applications such as self-driving, space exploration, urban air mobility, and human-robot collaboration. 
Planning, or sequential decision-making, in these settings requires a prediction model of the other agents, which can be generated through a game theoretic framework. 

Recently, the success of AlphaZero~\cite{Silver1140} at the game of Go has popularized a self-improving machine learning algorithm: bias a Monte Carlo Tree Search with value and policy neural networks, use the tree statistics to train the networks with supervised learning and then iterate over these two steps to improve the policy and value networks over time. However, this algorithm is designed for classical artificial intelligence tasks (e.g. chess or Go), and applications in multi-robot domains require different assumptions: continuous state-action, decentralized evaluation, partial information, and limited computational resources. 
To the best of our knowledge, our work is the first to provide a complete multi-robot adaption from algorithm design to hardware experiment of the AlphaZero method.

\begin{figure}
    \centering
    \includegraphics[width=0.80\linewidth]{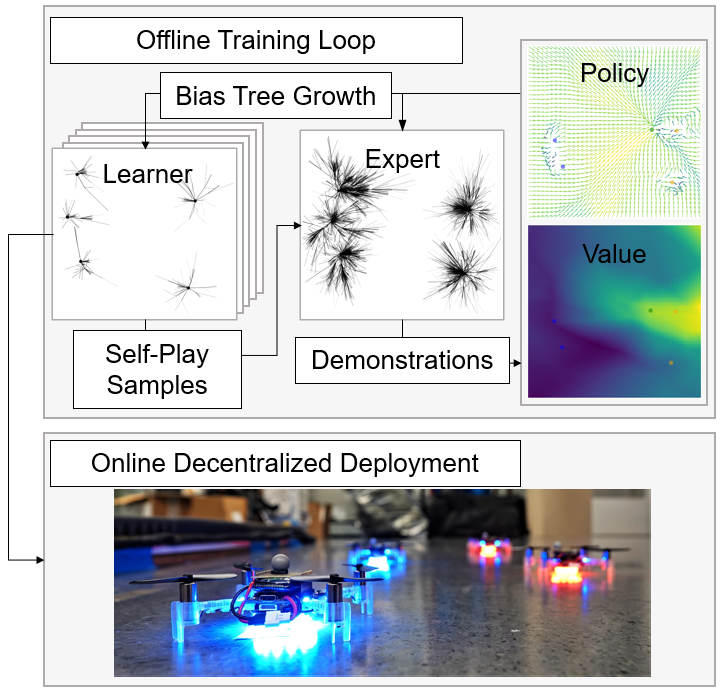}
    \caption{
    We propose an AlphaZero-like method for multi-robot applications such as the Reach-Target-Avoid game.
    Offline, the learner finds relevant states through self-play, for which the expert generates demonstration data. The data is used to train policy and value neural networks, which are used to bias both tree growths. 
    At runtime, the learner is deployed on each robot to generate an action with local information and little computation expense.}
    \label{fig:overview}
\end{figure}

The overview of our algorithm is shown in Fig.~\ref{fig:overview}. The key algorithmic innovation of our approach is to create two distinct MCTS policies to bridge the gap between high-performance simulation and real-world robotic application: the ``expert'' tree search is centralized and has access to perfect information and large computational resources, whereas the ``learner'' tree search is decentralized and has access to partial information and limited computational resources. During the offline phase, the neural networks are trained in an imitation learning style using the self-play states of the learner and the high-quality demonstrations of the expert. 
The expert's high-quality demonstrations enable policy improvement through iterations to incrementally improve the policy and value networks. 
The learner's self-play samples states that should appear more frequently at runtime. At deployment, each robot uses the learner to effectively plan online with partial information and limited computational budget. 
Our contribution is the Neural Tree Expansion algorithm that extends AlphaZero methods to (i) decentralized evaluation with local information, (ii) continuous state-action domain, and (iii) limited computational resources. 

We validate our method in simulation and experiment. We demonstrate numerically that our approach generates compact trees of similar or better performance with 20 times fewer nodes, and the resulting policies play the Reach-Target-Avoid differential game with double-integrator dynamics significantly better than the current state of the art~\cite{coonControlStrategiesMultiplayer2017}. 
Our method is compatible with arbitrary game specifications; we demonstrate this by generating visual examples of canonical games in Fig.~\ref{fig:NTE_general} and empirical evaluation of the Reach-Target-Avoid game for double-integrator and $3$D Dubin's vehicle dynamics in Sec.~\ref{sec:experiments}.
Our hardware experiments demonstrate that the solutions are robust to the gap between simulation and real world and neural expansion generates compact search trees that are effective real-time policies.


\textit{Related Work:} Our work relates to multiple communities: planning, machine learning, and game theory. 
Planning, or sequential decision-making, problems can be solved in an online setting with Monte Carlo Tree Search (MCTS)~\cite{Kochenderfer_2015}. 
MCTS searches through the large decision-making space by rolling out simulated trajectories and biasing the tree growth towards areas of high reward~\cite{Browne_2012}. 
MCTS was first popularized by the Upper Confidence Bound for Trees algorithm~\cite{Kocsis_2006} that uses a discrete-action, multi-armed bandit solution to balance exploration and exploitation in node selection. 
Recent work uses a non-stationary bandit analysis to propose a polynomial, rather than logarithmic, exploration term~\cite{Shah_2020}. 
As an anytime algorithm, the space and time complexity of MCTS is user-determined by the desired number of simulations. Recent finite sample complexity results of MCTS~\cite{Shah_2020,Mao_2020} show the error in root node value estimation converges at a rate of the order $n^{-1/2}$ where $n$ is the number of simulations.

Application of MCTS to a dynamically-constrained robot planning setting requires extending the theoretical foundations to a continuous state and action space. In general, the recent advances in this area answer two principal questions: i) how to select an action, and ii) when a node is fully expanded. 
Regarding the former question, some solutions select an action using the extension of the multi-armed bandit in continuous domains~\cite{Mansley_2011}, whereas our approach uses a policy network to generate actions. 
Regarding the latter question, a popular method to determine whether a node is expanded is to use progressive widening and variants; we adapt one such method, the Polynomial Upper Continuous Trees (PUCT) algorithm~\cite{Auger_2013}.  
Despite the advance in theory for continuous action spaces, there have been relatively few studies of biasing continuous MCTS with deep neural networks~\cite{Moerland_2018}. 

The key idea of AlphaZero~\cite{Silver1140} is using MCTS as a policy improvement operator; i.e. given a policy neural network to guide MCTS, the resulting search produces an action closer to the optimal solution than that generated by the neural network. Then, the neural network is trained with supervised learning to imitate the superior MCTS policy, matching the quality of the network to that of MCTS in the training domain. By iterating over these two steps, the model improves over time. The first theoretical analysis of this powerful method is recently shown for single-agent discrete action space problems~\cite{Shah_2020}. In comparison, our method is applied to a continuous state-action, multi-agent setting. Whereas AlphaZero methods use the policy network to bias the node selection process, i.e. given a list of actions, select the best one, our policy network is an action generator for the expansion process to create edges to children, i.e. given a state, generate an action. A neural expansion operator has previously been explored in motion planning~\cite{Chen_2020}, but not decision-making. In addition, our method's supervised learning step is closer to imitation learning, as used in DAgger~\cite{Ross_2011}, because the learner benefits from an adaptive dataset generation of using self-play to query from an expert. 

Although the AlphaZero methods use a form of supervised learning to train the networks, they can be classified as a reinforcement learning method because the networks are trained without a pre-existing labelled dataset. Policy gradient~\cite{Sutton_1999} is a conventional reinforcement learning solution and there are many recent advances in this area~\cite{Prajapat_2020}. 
Adding an underlying tree structure to deep reinforcement learning provides a higher degree of interpretability and a more stable learning process, enabled by MCTS's policy improvement property. 

In contrast to data-driven methods, traditional analytical solutions can be studied and derived through differential game theory. The game we study, Reach-Target-Avoid, was first introduced and solved for simple-motion, 1 vs. 1 systems~\cite{Isaacs_1965}. Later, multi-robot, single-integrator solutions have been proposed~\cite{garciaOptimalStrategiesClass2020,Yan_2019}. Solutions considering multi-robots with non-trivial dynamics, such as the double-integrator~\cite{coonControlStrategiesMultiplayer2017}, are an active area of research. Shepherding, herding, and perimeter defense are variants of the Reach-Target-Avoid game and are also active areas of research~\cite{paranjape2018robotic,Hu_2020, Nardi_2018, Shishika_2020}. 




\section{Problem Formulation}
\label{sec:problem_formulation}
\textit{Notation:} We denote the learning iteration with $k$ and the physical timestep with a subscript $t$, which is suppressed for notation simplicity, unless necessary. Robot-specific quantities are denoted with $i$ or $j$ superscript, and, in context, the absence of superscript denotes a joint-space quantity, e.g. the joint state vector is the vertical stack of all individual robot vectors, $\vs_t = [\vs_t^1;\hdots;\vs_t^N]$ where $N$ is the number of robots. 
\begin{definition}
    A partially observable stochastic game (POSG) is defined by a tuple: $\gG = \langle \sI,\sS,\sA,\fT,\fR,\sZ,\fO,\cH \rangle$ where: 
    $\sI = \{1,...,\cN\}$ is the set of robot indices, 
    $\sS$ is the set of joint robot states,
    $\sA$ is the set of joint robot actions,
    $\fT$ is the joint robot transition function where $\fT(\vs_t, \va_t, \vs_{t+1}) = \mathds{P}(\vs_{t+1} | \vs_t, \va_t)$ is the probability of transitioning from joint state $\vs_t$ to $\vs_{t+1}$ under joint action $\va_t$,
    $\fR$ is the set of joint robot rewards functions where $\fR^i(\vs, \va^i)$ is the immediate reward of robot $i$ for taking local action $\va^i$ in joint state $\vs$, 
    $\sZ$ is the set of joint robot observations, 
    $\fO$ is the set of joint robot observation probabilities where $\fO(\vz, \vs, \va) = \mathds{P}(\vz | \vs, \va)$ is the probability of observing joint observation $\vz$ conditioning on the joint state and action, and 
    $\cH$ is the planning horizon. These joint-quantities can be constructed from the robot-specific quantity. 
    The solution of a POSG is a sequence of actions that maximizes the expected reward over time and is often characterized by a policy or value function. 
\end{definition}
\begin{assumption}
    \label{assumption:deterministic}
   A deterministic transition function is assumed, thereby permitting rewriting $\fT$ with a dynamics function, $\ff$, as: 
    $\fT(\vs_t, \va_t, \vs_{t+1}) = \mathds{I}(\ff(\vs_t,\va_t) = \vs_{t+1})$
    where $\mathds{I}$ is an indicator function. Similarly, rewriting $\fO$ with a deterministic observation function results in $\vz^i = \fh^i(\vs)$ for each robot $i$. 
\end{assumption}
Assumption~\ref{assumption:deterministic} can be relaxed by considering specialized variants that are not the focus of this work. For example, realistic robotic scenarios with localization uncertainty from measurement noise can be handled with the observation widening variant~\cite{Sunberg_2018}.

\begin{figure}
    \centering
    \begin{subfigure}[m]{0.225\textwidth}
        \centering
        \includegraphics[width=\textwidth,height=4cm,keepaspectratio]{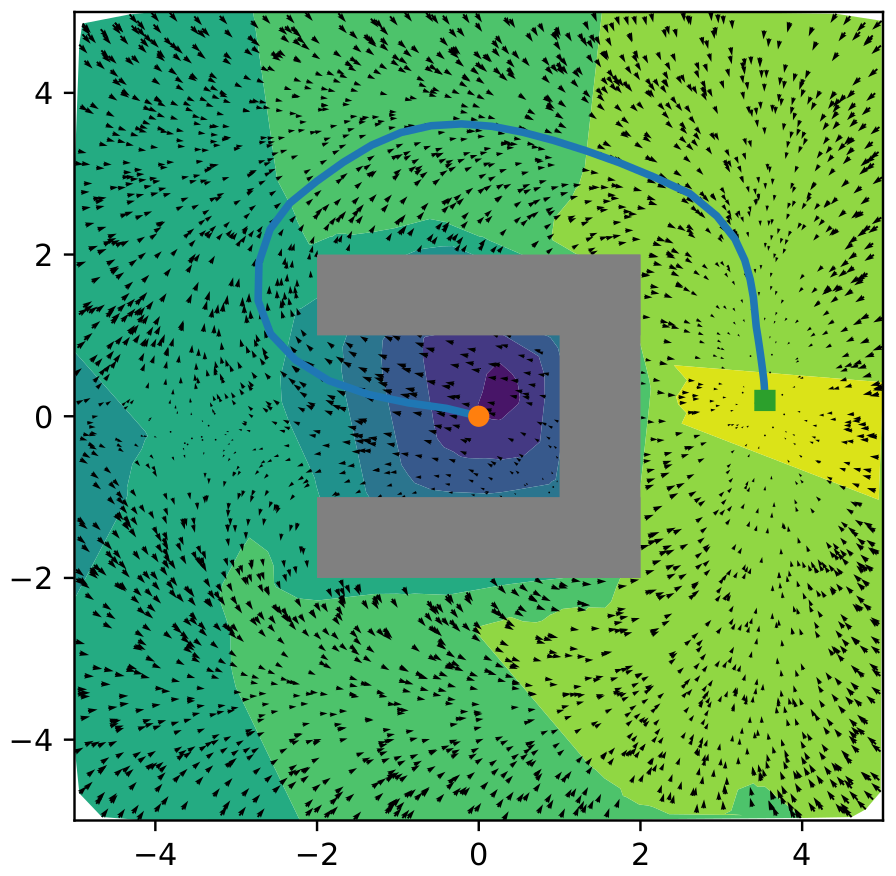}
        \caption{NTE finds the intuitive value and policy function for the ``bugtrap'' motion planning problem. The robot starts at the orange dot, and terminates at the green square after reaching the goal.}
    \end{subfigure}    
    \hspace{0.25cm}
    \begin{subfigure}[m]{0.225\textwidth}
        \centering
        \includegraphics[width=\textwidth, height=4cm,keepaspectratio]{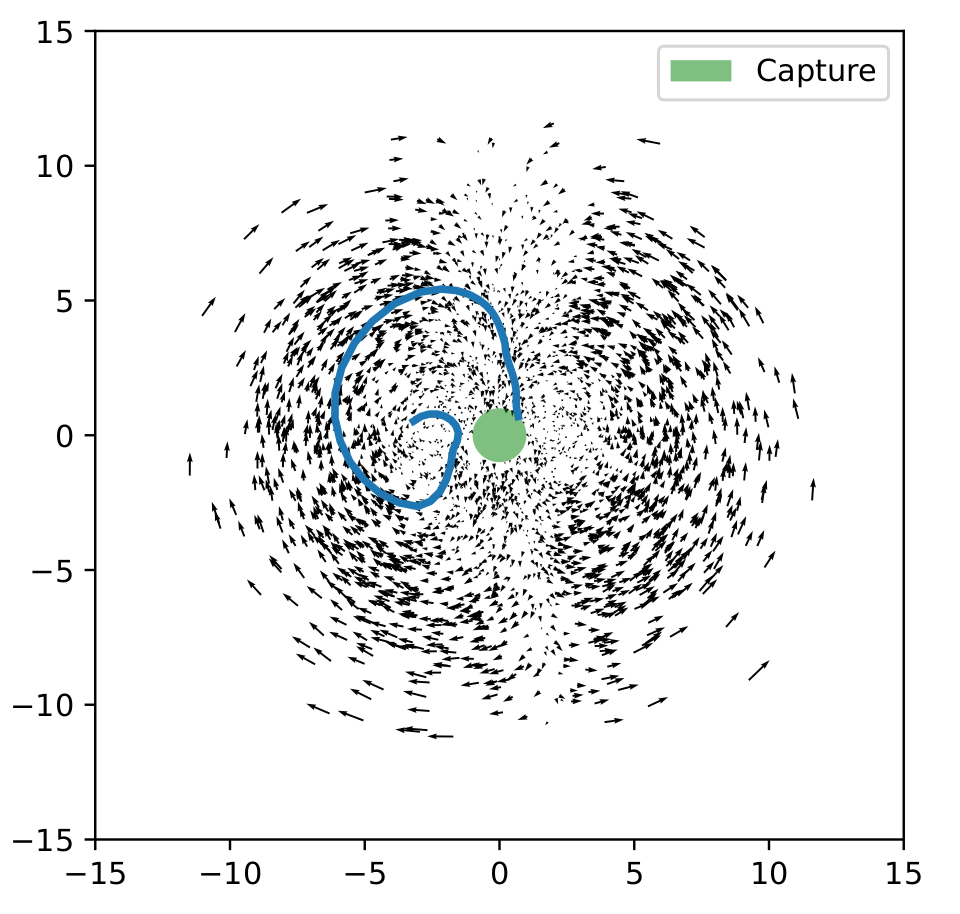}
        \caption{
        The state trajectories generated by NTE approximate the primary solution and barrier surface for the ``homicidal chauffeur'' game~\cite{Isaacs_1965}. The plot is shown in Isaac's reduced space with an example trajectory in blue terminating in a capture condition.
        }
    \end{subfigure}
    \hspace{0.25cm}
    \begin{subfigure}[t]{0.225\textwidth}
        \centering
        \includegraphics[width=\textwidth,height=4cm,keepaspectratio]{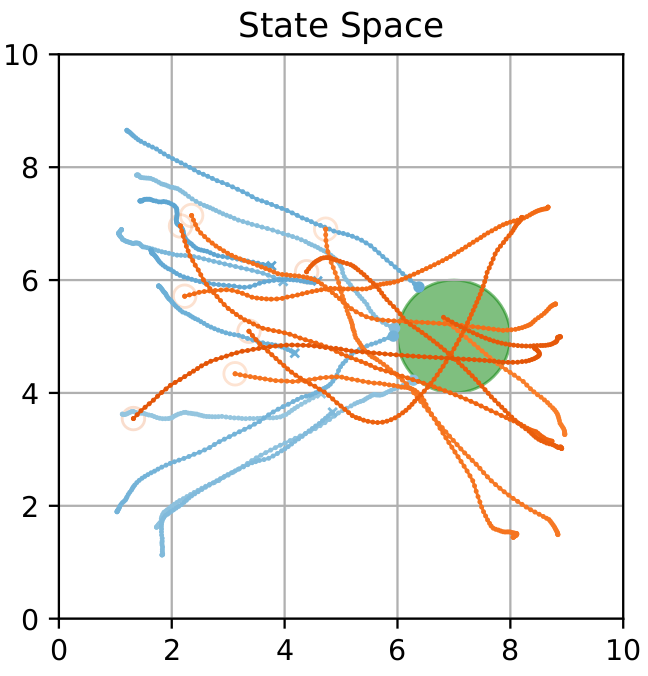}
        \caption{NTE scales to high dimensional team games for the $10$ agent vs. $10$ agent ``Reach-Target-Avoid'' with double-integrator dynamics. }
        \label{fig:example_state_space}
    \end{subfigure}
    \hspace{0.25cm}
    \begin{subfigure}[t]{0.225\textwidth}
        \centering
        \includegraphics[width=\textwidth,height=4cm]{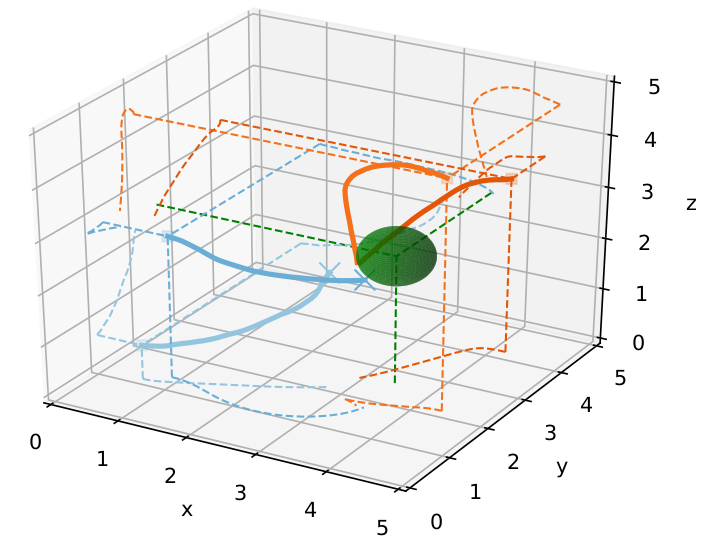}
        \caption{
        NTE is compatible with arbitrary dynamics, here is the ``Reach-Target-Avoid'' game with $3$D Dubin's vehicle dynamics. 
        }
        \label{fig:dubins3d_state_space_ex}
    \end{subfigure} 
    \caption{Neural Tree Expansion (NTE) can be applied to decision-making problems such as single agent motion planning, canonical differential games, and high-dimensional team games.}
    \label{fig:NTE_general}
\end{figure}

\textit{Problem Statement:} At time $t$, each robot $i$ makes a local observation, $\vz^i$, uses it to formulate an action, $\va^i$, and updates its state, $\vs^i$, according to the dynamical model. Our goal is to find policies for each robot, $\fpi^i : \sZ^i \rightarrow \sA^i$ that synthesizes actions from local observations through: 
\begin{align}
	\vz^i_t &= \fh^i(\vs_t), \ \ 
	\va^i_t = \fpi^i(\vz^i_t),
\end{align}
to approximate the solution to the general-sum, game theoretic optimization problem: 
\begin{equation}
\begin{aligned}
\label{eq:dec_decision_making}
\va^{i^*}_t &= \argmax_{\{ \va^i_\tau | \forall \tau\}} \sum_{\tau = t}^{t + \cH} \fR^i(\vs_\tau,\va_\tau^i) \quad 
\textrm{ s.t.} \\
	\vs_{\tau+1} &= \ff(\vs_\tau,\va_\tau), \ 
	\vs^i_\tau \in \sX^i, \ 
    \va^i_\tau \in \sU^i,  \ 
    \vs^i_{\tau_0} = \vs^i_0, \ \forall i,\tau 
\end{aligned}
\end{equation}
where $\sU^i \subseteq \sA^i$ is the set of available actions (e.g. bounded control authority constraints), and $\sX^i \subseteq \sS^i$ is the set of safe states (e.g. collision avoidance) and $\vs^i_0$ is the initial state condition. The optimization problems for each robot $i$ are simultaneously coupled through the evolution of the global state vector $\vs$, where each robot attempts to maximize its own reward function $\fR^i$. We evaluate our method on canonical cases and present visualizations in Fig.~\ref{fig:NTE_general}. 

\textit{Reach-Target-Avoid Game:} An instance of the above formulation is the Reach-Target-Avoid game~\cite{Isaacs_1965} for two teams of robots, where team $A$ gets points for robots that reach the goal region, and team $B$ gets points for defending the goal by tagging the invading robots first. The teams are parameterized by index sets $\sI_A$ and $\sI_B$, respectively, where the union of the two teams represents all robots, $\sI_A \ \cup \ \sI_B = \sI$. An example of the Reach-Target-Avoid game is shown in Fig.~\ref{fig:example_state_space}, where the red robots try to tag the blue robots before the blue robots reach the green goal region. The $x$ and $o$ on the trajectory indicates tagged state and reached goal. 

\textit{Dynamics:} We consider the discrete-time double-integrator system for the $i\th$ robot as a motivating example: 
\begin{align}
    \label{eq:di_dynamics}
    \vs^i_{t+1} &= 
    \begin{bmatrix} 
        \vp^i_t \\ 
        \vv^i_t 
    \end{bmatrix} + 
    \begin{bmatrix} 
        \vv^i_t \\ 
        \va^i_t 
    \end{bmatrix} \cdelta_t
\end{align}
where $\vp^i$ and $\vv^i$ denote position and velocity and $\cdelta_t$ denotes the simulation timestep. We use a simultaneous turn game formulation where at a given timestep, each team's action is chosen without knowledge of the other team's action. 

\textit{Admissible State and Action Space:} The admissible state space for each robot is defined by the following constraints: remain inside the position and velocity bounds, $\cpul$ and $\cvul$, and avoid collisions within the physical robot radius, $\cphysicalradius$:
\begin{align}
    \label{eq:environment_bounds}
    &\| \vp^i \|_\infty \leq \cpul, \ \| \vv^i \|_2 \leq \cvul, \ \ 
    \| \vp^j - \vp^i \| > \cphysicalradius, \ \forall i,j \in \sI 
\end{align}
For each robot $i$ on team $A$ ($\forall i \in \sI_A$), the admissible state space has an additional constraint: avoid the robots on team $B$ by at least the tag radius:
\begin{align}
    \label{eq:defender_tag_and_goal_tag}
    &\|  \vp^i-\vp^j  \| > \ctagradius, \ 
    \ \forall j \in \sI_B
\end{align}
Then, the admissible state space for each team can be written compactly, e.g. 
$ \sX^i = \{ \vs^i \in \sS^i \ | \ \text{ s.t. \eqref{eq:environment_bounds}, \eqref{eq:defender_tag_and_goal_tag}} \}, \ \  \forall i \in \sI_A $. 
The admissible action space for each robot is constrained by its acceleration limit:
$ \sU^i = \{ \va^i \in \sA^i \ | \ \| \va^i \|_2 \leq \caul \}, \ \forall i \in \sI$, where $\caul$ is the robot's acceleration limit. 
When a robot exits the admissible state or action space, or a robot on team $A$'s position is within an $\cgoalradius$ radius about the goal position $\vp_g$, it becomes inactive. 

\textit{Observation Model:} Under Assumption~\ref{assumption:deterministic}, for each robot $i$, we define a measurement model that is similar to visual relative navigation $\fh^i: \sS \rightarrow \sZ^i$, which measures the relative state measurement between neighboring robots, as well as relative state to the goal. Specifically, an observation is defined as:
\begin{align}
	\vz^i &= \left[ \vg - \vs^i, \{\vs^j - \vs^i\}_{j \in \sN^i_A}, \{ \vs^j - \vs^i \}_{j \in \sN^i_B} \right], \label{eq:observation}
\end{align}
where $\vg$ is the goal position embedded in the state space, e.g. for 2D double-integrator $\vg = [\vp_g;0;0]$. Then, $\sN^i_A$ and $\sN^i_B$ denote the $i\th$ robot's neighbors on team $A$ and $B$, respectively. These sets are defined by each robot's sensing radius, $\csensingradius$:  
\begin{equation}
	\sN^i_A = \{ j \in \sI_A \ | \ \| \vp^j - \vp^i \|_2 \leq \csensingradius \}.
\end{equation}

\textit{Reward:} The robot behavior is driven by the reward function; the inter-team cooperation behavior is incentivized by sharing the reward function and intra-team adversarial behavior is incentivized by assigning complementary reward functions only dependent on global state, $\fR^i(\vs) = -\fR^j(\vs), \ \forall i \in \sI_A, j \in \sI_B$. The reward can be defined by a single, robot-agnostic game reward, $\fR(\vs)$ that team $A$ tries to maximize and team $B$ tries to minimize. The game reward is $0$ until the terminal state where the game reward and traditional value function are identical and defined as:
\begin{alignat}{2}
    \fV(\vs_{t}) &= \sum_{\forall i \in \sI_A} \mathds{I} (\|\vp^i_t - \vp_g \|_2 \leq \cgoalradius),
\end{alignat}
i.e. the value is the number of team $A$ robots in the goal region. The indicator function payoff is known to be sparse and makes traditional search and reinforcement learning techniques ineffective~\cite{Florensa_2017}. The game termination occurs when all robots on team $A$ are inactive; typically when they have reached the goal or been tagged by a robot on team $B$. 

\section{NTE Algorithm Description}
\label{sec:algorithm}
We present the meta-algorithm, the expert and learner NTE, and the policy and value neural networks. 


\subsection{Meta Self-Improving Algorithm}
The input of the meta-learning is the POSG game described in Sec.~\ref{sec:problem_formulation}, and the outputs are the policy and value neural networks, $\fhpi$ and $\fhv$. The goal of the meta-learning is to improve the models across learning iterations, especially in relevant state domains, such that at runtime, the robots can evaluate the learner. The desired model improvement can be expressed by decreasing some general probability distribution distance between the policy network and the optimal policy function: 
\begin{align}
    \label{eq:meta_learning_goal}
    \text{dist}(\fhpi_{k}(\vz), \pi^*(\vs)) \leq \text{dist}(\fhpi_{m}(\vz), \pi^*(\vs)), \ \forall k > m , \ \forall{\vs}
\end{align}
where $\fpioptimal$ is the unknown optimal policy function that inputs a joint state and returns a joint action, $\fhpi_{k}$ is the policy network we train, and $k,m$ are learning iteration indices. Specifically, we train robot-specific policies ${\fhpi}^i_k$ that map local observation to local action and compose them together to create the joint policy $\fhpi_{k} = [\fhpi^1_k;\hdots;\fhpi^{|\sI|}_k]$ that maps joint observation, $\vz$, to joint action, $\va$. Adapting the proof concept in~\cite{Shah_2020} to our setting, the policy improvement can be shown by validating two properties and then iterating: (i)~bootstrap
\begin{align}
    \label{eq:bootstrap} 
    &\text{dist}(\fpioptimal(\vs),\fepi_k(\vs,\fhpi_{k},\fhv_{k})) \leq \text{dist}(\fpioptimal(\vs),\fhpi_{k}(\vz)), \  \forall{\vs} 
\end{align}
and, after generating an appropriate dataset, (ii)~learning
\begin{align}
    &\sD_{\fpi} = \{\vs, \va\} \text{ with } \va = \fepi_k(\vs,\fhpi_{k},\fhv_{k}) \\ 
    \label{eq:learning}
    &\text{dist}(\fhpi_{k+1}(\vz),\fpioptimal(\vs)) \approx \text{dist}(\va,\fpioptimal(\vs)) , \ \forall (\vs,\va) \in \sD_{\fpi}
\end{align}
where $\fhv_k$ is the value network and $\fepi$ is the expert that maps state to joint action while biased by the policy and value neural networks. Intuitively, the bootstrap property of MCTS~\eqref{eq:bootstrap} generates a dataset of policy samples superior to that of the policy network, and then the supervised learning property~\eqref{eq:learning} matches the quality of the policy network to the quality of the new dataset. 

Validating these two properties drives the design of our meta-learning algorithm in Algorithm~\ref{algo:overview}. 
At each learning iteration $k$, each robot's policy network is trained in the following manner: a set of states is generated from self-play of the multi-agent learners, $\flpi_k$ and then the expert, $\fepi_k$, searches on these states to create a dataset of learning targets for the supervised learning. 
The value network, $\fhv_k$ is trained to predict the outcome of the game if each team were to play with the joint policy network $\fhpi_k$. 
Because the centralized expert has perfect information, coordination and large computational resources, the bootstrap property is more likely to hold. Furthermore, as the learner generates state space samples through self-play, the dataset is dense in frequently visited areas of the state space, and the models will be more accurate there, validating the learning property. Finally, we specify a simple POSG generator in Line~\ref{eq:curriculum} of Algorithm~\ref{algo:overview} to select opponent policies and game parameters for self-play. 

\begin{algorithm}
\caption{Meta Self-Improving Learning}
\label{algo:overview}
\SetAlgoLined
\DontPrintSemicolon
\SetKwProg{Def}{def}{:}{}
\Def{Meta-Learning($\gG^*$)}{
    $\fhpi_0, \fhv_0 = \text{None}, \text{None}$ \; 
    \For{$ k = 0, \hdots, K $}{
        $\{\gG\}_k = \mathrm{makePOSG}(\gG^*, k)$ \;
        \label{eq:curriculum}
        \For{$i \in \sI$}{ 
            \label{eq:robot_loop}
            $\{\vs\} = \mathrm{selfPlay}(\flpi((.),\fhpi_k,\fhv_k),\{\gG\}_k)$ \;
            \tcc{$\mathrm{Search}$ from Algorithm 2}
            $\sD^i_\fpi = \{\vs,\fepi_k.\mathrm{Search}(\vs,\fhpi_k,\fhv_k))\} $ \; 
            \label{eq:policy_dataset}
            ${\fhpi_{k+1}}^i = \mathrm{trainPolicy}(\sD^i_{\fpi})$ \;
            \label{eq:policy_training}
        }
        $\sD_\fV = \{ \vs, \text{selfPlay}(\fhpi_k,\{\gG\}_k)\} $ \; 
        \label{eq:value_dataset}
        $\fhv_{k+1} = \mathrm{trainValue}(\sD_\fV)$ \;
        \label{eq:value_training}
    }
}
\end{algorithm}

\subsection{Neural Tree Expansion}
In order to specify the expert and learner policies, we first explain their common search tree algorithm shown in Algorithm~\ref{algo:bmcts} and adapted from~\cite{Browne_2012} to our setting. For a complete treatment of MCTS, we refer the reader to~\cite{Browne_2012}. 

The biased MCTS algorithm begins at some start state $\vs$ and grows the tree until its computational budget is exhausted, typically measured by the number of nodes in the tree, $L$. 
Each node in the tree is a state, $\vs$, each edge is an action $\va$, and each child is the new state after propagating the dynamics. 
Each node in the tree, $\onode$, is initialized with a state vector and an action edge to its parent node, $\onode^p$, i.e. $\onode = \textit{Node}(\vs,(\onode^p,\va))$. Each node stores the state vector, $S(\onode)$, the number of visits to the node, $N(\onode)$, its children set, $C(\onode)$, and its action set, $A(\onode,\onode'), \ \forall \onode' \in C(\onode)$. The growth iteration in the main function, \textit{Search}, has four steps: (i) node selection, \textit{Select}, selects a node to balance exploration of space and exploitation of rewards (ii) node expansion, \textit{Expand}, creates a child node by forward propagating the selected node with an action either constructed by the neural network or by random sampling, (iii) \textit{DefaultPolicy} collects terminal reward statistics by either sampling the value neural network or by rolling out a simulated state trajectory from the new node, and (iv) \textit{Backpropagate} updates the number of visits and cumulative reward up the tree. The action returned by the search is the child of the root node with the most visits. 
The primary changes we make from standard MCTS are the integration of neural networks, highlighted in Algorithm~\ref{algo:bmcts}.

The behavior of other agents is modelled in a turn-based fashion: each depth in the tree corresponds to the turn of an agent and their action is predicted by selecting the best node for their cost function. Intuitively, the MCTS search plans for all robots, assuming that they maximize their incentive. MCTS is known to converge to the minimax tree solution~\cite{Browne_2012}. 




\begin{algorithm}
\caption{Neural Tree Expansion}
\label{algo:bmcts}
    \DontPrintSemicolon
    \tcc{$\fhpi,\fhv$ from Algorithm 1}
    \SetKwProg{Def}{def}{:}{}
    \Def{Search($\vs,\fhpi,\fhv$)}{
        $\onode_0 \leftarrow  \textit{Node}(\vs,\text{None})$ \;
        \For{$l=1,\hdots,L$}{
            $\onode_l \leftarrow \textit{Expand}(\textit{Select}(\onode_0,\fhpi))$ \;
            $v \leftarrow \textit{DefaultPolicy}(S(\onode_l))$ \; 
            $\textit{Backpropagate}(\onode_l,v)$ \; 
        }
        return $A(\onode_0,\argmax_{\onode' \in C(\onode_0)} N(\onode')) $\;
    }   
    \Def{Expand($\onode,\fhpi$)}{
        \tikzmk{A} $\alpha \sim \mathds{U}(0,1)$ \; 
        \label{eq:tree_policy_heuristic_start}h
        \If{$\alpha < \beta_\fpi$}{
            $\va \leftarrow [\va^1,\hdots,\va^{|\sI|}], \ \ \va^i \sim {\fhpi}^i(\fh^i(\vs)), \ \  \forall i \in \sI$
            \label{eq:tree_policy_heuristic_stop}\markline{-25pt}\;
        }\Else{
            $\va \leftarrow [\va^1,\hdots,\va^{|\sI|}], \ \ \va^i \sim \sU^i, \ \  \forall i \in \sI$ \;
        }
        $\onode' \leftarrow \textit{Node}(\ff(\vs,\va),(\onode,\va))$ \;
        return $\onode'$ \; 
    }
    \Def{DefaultPolicy($\vs,\fhv$)}{
        \tikzmk{A} $\alpha \sim \mathds{U}(0,1)$ \; 
        \label{eq:tree_value_heuristic_start}
        \If{$\alpha < \beta_\fV$}{
            $v \sim \fhv(\fh_y(\vs))$
            \label{eq:tree_value_heuristic_stop}\markline{-25pt}\;
        }\Else{
            \While{$\vs$ \text{ is not terminal }}{
                $\va \leftarrow [\va^1,\hdots,\va^{|\sI|}], \ \ \va^i \sim \sU^i, \ \  \forall i \in \sI$ \;
                $\vs \leftarrow \ff(\vs,\va)$ \; 
            }
            $v \leftarrow \fV(\vs)$ \;
        }
        return $v$ \;
    }
\end{algorithm}

\subsection{Expert NTE}
The expert, $\fepi$, is a function from joint state $\vs$ to joint action $\va$. The expert computes the action by calling $\textit{Search}$ in Algorithm~\ref{algo:bmcts} with a large number of nodes $L_\text{expert}$. 
The expert produces a coordinated team action by selecting the appropriate indices of the joint-space action, where the remaining, unused actions represent the predicted opponent team action. 
The expert's perfect information, centralized response, and large computational budget is necessary to guarantee the bootstrap property~\eqref{eq:bootstrap}. 
We found that if the expert is given less computational resources, the learning process is not stable and the quality of the policy and value networks deteriorates over learning iterations. Many of the desirable properties of the expert for theoretical performance make it an infeasible solution for multi-robot applications, motivating the design of the learner. 

\subsection{Learner NTE} 
The learner for robot $i$, $\flpi$, is a function from local observation $\vz^i$ to local action $\va^i$. The learner computes the action by reconstructing the state from its local observation naively: $\tilde{\vs}(\vz) = \{ \tilde{\vs}^j \}, \ \forall j \in \sN^i_A \cup \sN^i_B$, 
where we assume that the learner has prior knowledge of the absolute goal location.
Then, the learner calls $\textit{Search}$ in Algorithm~\ref{algo:bmcts} with the estimated state and a small number of nodes $L_\text{learner}$, $L_\text{learner} < L_\text{expert}$. The final action $\va^i$ is selected from the appropriate index of the joint-space action returned by $\textit{Search}$. 
Because the learner only selects a single action from the joint-space action, the learner is predicting the behavior of robots on both teams. 
This communication-less implicit coordination enables operation in bandwidth-limited or communication-denied environments. 

\subsection{Policy and Value Neural Networks}
\label{sec:heuristics}
We introduce each neural network with its dataset generation and training in Algorithm~\ref{algo:overview}, and its effect on tree growth via integration into Algorithm~\ref{algo:bmcts}. 

\textit{Policy Network: } 
The policy network for robot $i$ maps observations to the action distribution for a single robot and is used to create children nodes. 
The desired behavior of the policy network is to generate individual robot actions with a high probability of being near-optimal expansions given the current observation, i.e. generate edges to nodes with a high number of visits in the expert search. 

The dataset for each robot $i$'s policy network is composed of observation action pairs as computed in Line~\ref{eq:policy_dataset} of Algorithm~\ref{algo:overview}. 
The action label is calculated by querying the expert at some state, extracting the root node's child distribution, and calculating the action label as the first moment of the action distribution, weighted by the relative number of visits: 
\begin{align}
    \va^i_l &= \sum_{\onode'\in C(\onode_0)} \frac{N(\onode')}{N(\onode_0)} A^i(\onode_0,\onode')  
\end{align}
where $\va^i_l$ is the action label and $\onode_0$ is the root node. Recall that $C(\cdot)$ is a node's set of child nodes, $N(\cdot)$ is the number of visits to a node, and $A^i(\onode_0,\onode')$ is the $i$th robot's action from root node $\onode_0$ to child node $\onode'$. 
Next, we change the input from state to observation by applying robot $i$'s observation model, $\vz^i_l = h^i(\vs_l)$. This is a global-to-local learning technique to automatically synthesize local policies from centralized examples~\cite{Riviere_2020}. The collection of observation-action samples can be written in a dataset as $\sD^i_\fpi = \{ (\vz^i_l , \va^i_l) | \ l = 1,\hdots \}$. 

The policy network training in Line~\ref{eq:policy_training} in Algorithm~\ref{algo:overview} is cast as a multivariate Gaussian learning problem, i.e. the output of the neural network is a mean, $\vmu$ and variance $\vSigma$. An action sample $\hat{\va}^i_l \sim \flpi(\vz^i_l)$ can then be computed by sampling $\epsilon$ and transforming it by the neural network output: 
\begin{align}
    \label{eq:policy_query}
    \hat{\va}^i_l = \vmu(\vz^i_l) + \vSigma(\vz^i_l) \epsilon, \ \ 
    \epsilon \sim \mathcal{N}(0,I)
\end{align}
The input, $\vz^i_l$, is encoded with a DeepSet~\cite{Zaheer_2017} feedforward architecture similar to~\cite{Riviere_2020} that is compatible with a variable number of neighboring robots. 
The maximum likelihood solution to the multivariate Gaussian problem is found by minimizing the following loss function: 
\begin{align}
    \label{eq:policy_loss}
    &\fL = \mathds{E} \sum_l (\va^i_l - \vmu)^T \vSigma^{-1} (\va^i_l - \vmu) + \frac{1}{2} \mathrm{ln} |\vSigma| \\ 
    &{\fhpi}^i = \argmin_{\fhpi^i \in \Pi^i} \mathds{E} \  \fL(\vmu(\vz^i_l),\vSigma(\vz^i_l),\va^i_l)
\end{align}
where $\vmu$ and $\vSigma$ are generated by the neural network given $\vz^i_l$, and $\va^i_l$ is the target. 

The policy neural network, $\fhpi^i$, is integrated in Line~\ref{eq:tree_policy_heuristic_start}--\ref{eq:tree_policy_heuristic_stop} in Algorithm~\ref{algo:bmcts} in the expansion operation by constructing a joint-space action from decentralized evaluations of the policy network for all the agents, and then forward propagating that action. We found that using a neural expansion, rather than neural selection as in AlphaZero, is necessary for planning with a small number of nodes in environments with many robots. 
For example, in a $10$ vs. $10$ game such as that shown in Fig.~\ref{fig:example_state_space}, the probability of sampling a control action from a uniform distribution that steers each robot towards the goal within $90$ degrees is $(1/4)^{10}$. 
If the learner policy is evaluated with standard parameters (see Sec.~\ref{sec:implementation}), it will generate 5 children, which collectively are not likely to contain the desired joint action. Using the neural network expansion operator will overcome this limitation by immediately generating promising child nodes. 
Similar to~\cite{Ichter_2018}, the expand operation switches between uniform random and neural network sampling at relative frequency $\beta_\fpi \in [0,1]$. The stochastic nature of both expansion modes enables the tree to maintain exploration.

\textit{Value Network}: The value network is used to gather reward statistics in place of a policy rollout, and is called in the \textit{DefaultPolicy} in Lines~\ref{eq:tree_value_heuristic_start}--\ref{eq:tree_value_heuristic_stop} of Algorithm~\ref{algo:bmcts}. The value network uses an alternative state representation to be compatible with the estimated state for local computation: 
\begin{align}
    \label{eq:alternative_observation}
    \vy &= \fh_y(\vs) = \left[ \{ \vs^j - \vg \}_{j\in\sN^i_A}, \{\vs^j - \vg \}_{j \in \sN^i_B}, n_{rg} \right] 
\end{align}
where $\fh_y(\vs)$ is the alternative observation function and $n_{rg}$ is the number of robots that have already reached the goal. The value network maps this alternative state representation to the parameters of a multi-variate Gaussian distribution.
The desired behavior of the value network is to predict the outcome of games if they were rolled out with the current policy network. The value function implementation in \textit{DefaultPolicy} is the same as AlphaZero methods. 

The value network dataset in Line~\ref{eq:value_dataset} of Algorithm~\ref{algo:overview} is generated for all robots at the same time and is composed of alternative state-value pairs. The $\vy_l$ state can be generated from $\vs$, and the value label, $v_l$ is generated by self-play with the current policy network. The dataset, $\sD_\fV$ can then be written as $\sD_\fV = \{ (\vy_l,v_l) | \ \forall l=1,\hdots \}$. 
Because AlphaZero methods use a policy selector rather than a generator, the dataset for the value network has to be made by rolling out entire games with MCTS. Instead, we generate the dataset by rolling out the policy network, which is much faster per sample, resulting in less total training time. 

The value network is trained in Line~\ref{eq:value_training} of Algorithm~\ref{algo:overview} with a similar loss function~\eqref{eq:policy_loss} as the policy network, using a learning target of the value labels, $v_l$, instead of the action $\va^i_l$. The value is also queried from the neural network in a similar fashion~\eqref{eq:policy_query}. The value network uses a similar model architecture as the policy network, permitting variable input size of $\vy$. Integration of the value network in \textit{DefaultPolicy} uses the same probabilistic scheme as the policy network with parameter $\beta_\fV$. 


\section{Experimental Validation}
\label{sec:experiments}

\subsection{MCTS and Learning Implementation}
\label{sec:implementation}
We implement Algorithm~\ref{algo:overview} in Python and Algorithm~\ref{algo:bmcts} in C$++$ with Python bindings. For the meta-algorithm, we only train the inner robot loop in Line~\ref{eq:robot_loop} of Algorithm~\ref{algo:overview} once per team because we use homogeneous robots and policies. Our MCTS variant uses the following hyperparameters: $L_\text{expert} = 10\,000$, $L_\text{learner} = 500$, $C_p = 2.0$, $C_{pw} = 1.0$, $\alpha_{pw} = 0.25$ and $\alpha_d = (1 - 3/(100 - 10d))/20$ where $d$ is the depth of the node. The neural frequency hyperparameters are $\beta_\fpi = \beta_\fV = 0.5$.  
The double-integrator game parameters are chosen to match the hardware used in the physical experiments (see Sec.~\ref{sec:hardware}). We use position bounds $\cpul = 1,2,3$ \SI{}{m} and constant velocity $\cvul = $ \SI{1.0}{m/s} and acceleration $\caul = $ \SI{2.0}{m/s^2} bounds. The tag, collision, and sensing radii are: $\ctagradius = $ \SI{0.2}{m}, $\cphysicalradius = $ \SI{0.1}{m}, $\csensingradius = $ \SI{2.0}{m}. We train for up to $5$ agents on each team. We use a simulation and planning timestep of $\cdelta =$ \SI{0.1}{s}. Each team starts at opposite sides of the environment, and the goal is placed closer to the defenders' starting position. 

We implement the machine learning components in PyTorch~\cite{pyTorch}. The datasets are of size \num{80000} points per iteration for both value and policy datasets. The meta learning algorithm is trained until convergence. 
The policy and value network models both use DeepSet~\cite{Zaheer_2017} neural network architecture, (e.g.~\cite{Riviere_2020}) where the inner and outer networks each have one hidden layer with $16$ neurons and appropriate input and output dimensions. 
All networks are of feedforward structure with ReLU activation functions, batch size of \num{1028}, and are trained over $300$ epochs.  

\subsection{Variants and Baseline}
In order to evaluate our method, we test the multiple learners and expert policies, each equipped with networks after $k$ learning iterations. To isolate the effect of the neural expansion, we consider the $k=0$ case for both learner and expert as an unbiased MCTS baseline solution. 

As an additional baseline for the double-integrator game, we use the solution from~\cite{coonControlStrategiesMultiplayer2017}. Their work adapts the exact differential game solution for simple-motion and single-robot teams proposed in~\cite{Isaacs_1965} to a double-integrator, multi-robot team setting. 
However, their adapted solution is not exact because it assumes a constant acceleration magnitude input and relies on composition of pair-wise matching strategies.  

\subsection{Simulation Results}
\label{sec:double_interator_evaluation}
We evaluate our expert and learner by initializing $100$ different initial conditions of a $3$ attacker, $2$ defender game in a \SI{3}{m} space. 
Then, we rollout every combination of variants, learning iterations, and baseline for both team $A$ and team $B$ policies, for a total of \num{12 100} games. For a single game, the performance criteria for team $A$ policies is the terminal reward and, in order to have consistency of plots (higher is better), the performance criteria for team $B$ policies is one minus the terminal reward.  
An example game with a different number of agents and environment size is shown in Fig.~\ref{fig:example_state_space} and its animation is provided in the supplemental video. The $10$ vs. $10$ game illustrates the natural scalability in number of agents of the decentralized approach and the generalizability of the neural networks, as they were only trained with data containing up to $5$ robots per team. 


\begin{figure}
    \centering 
    \begin{subfigure}[b]{0.4\textwidth}
        \centering
        \includegraphics[width=1.0\linewidth]{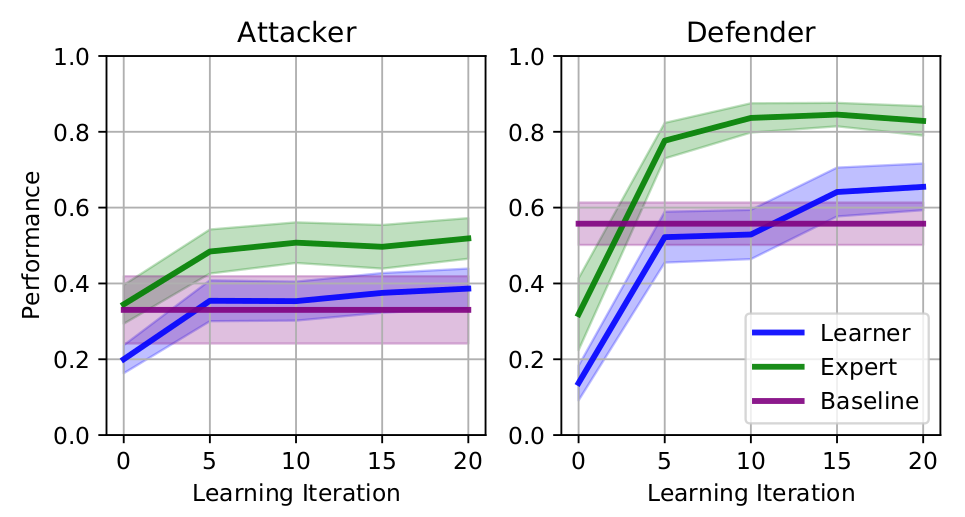}
        \caption{Double-integrator game evaluation: the thick lines indicate the average performance and the shaded area is the variance over 100 games. 
        }
        \label{fig:exp3_double_integrator}
    \end{subfigure}     
    \begin{subfigure}[t]{0.195\textwidth}
        \centering
        ~~~\includegraphics[width=0.92\textwidth]{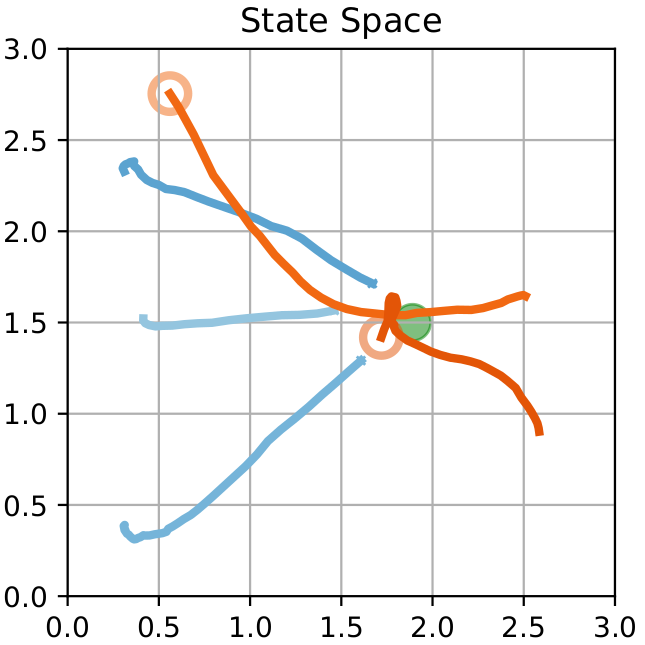}
        \caption{Learner defense (orange) finds emergent cooperative strategies to defend the goal (green).}
        \label{fig:exp3_double_integrator_learner}
    \end{subfigure}        
    \begin{subfigure}[t]{0.195\textwidth}
        \centering
        \includegraphics[width=0.92\textwidth]{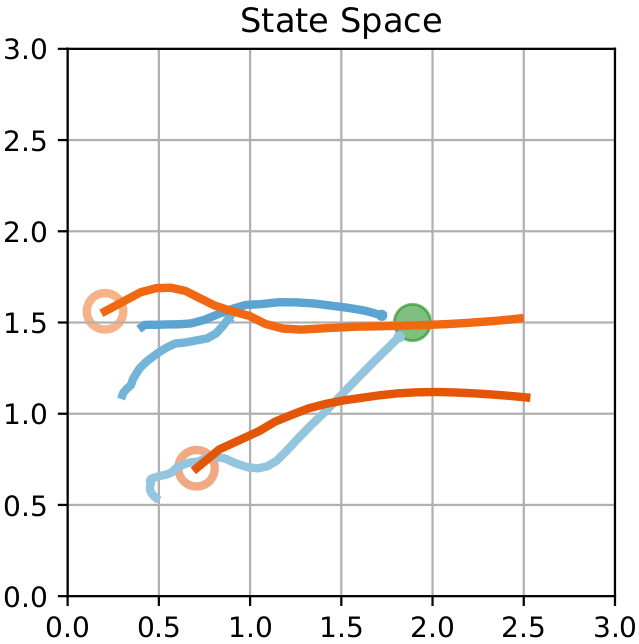}
        \caption{Baseline defense (orange) is vulnerable to learner's offensive (blue) dodge maneuver.}
        \label{fig:exp3_double_integrator_baseline}
    \end{subfigure}
    \caption{Double-integrator performance and strategy examples.} 
\end{figure}

The statistical results of the $3$ vs. $2$ experiment are shown in Fig.~\ref{fig:exp3_double_integrator} where the thick lines denote the average performance value and the shade is the performance variance. We find the expected results; for both team $A$ and team $B$, the learner with no bias has the worst performance, and learner with fully trained networks surpasses the centralized and expensive unbiased expert and approaches the biased expert. 
The baseline attacker is about the same strength as the unbiased expert, whereas the baseline defender is much stronger than the unbiased expert. 
In both cases, the fully-trained biased expert and learner are able to significantly outperform the baseline. 

To investigate the qualitative advantages of our method, we looked at the games where our learner defense outperformed the baseline defense and found two principal advantages: first, the learner defense sometimes demonstrated emergent coordination that is more effective than a pairwise matching strategy, e.g. one defender goes quickly to the goal to protect against greedy attacks while the other defender slowly approaches the goal to maintain its maneuverability, see Fig.~\ref{fig:exp3_double_integrator_learner}. Second, the learner attacker is sometimes able to exploit the momentum of the baseline defender and perform a dodge maneuver, e.g. the bottom left interaction in Fig.~\ref{fig:exp3_double_integrator_baseline}, whereas the learner defense is robust to this behavior. 
These examples show the learner networks can generate sophisticated, effective maneuvers.

As an additional experiment, we evaluate the learner (without retraining) in an environment with static and dynamic obstacles for $100$ different initial conditions; an example is shown in the supplementary video. In this environment, the fully trained learner outperforms the unbiased learner $0.246 \pm 0.022$, this value is calculated by summing the performance criteria difference across attacking and defending policies. This result demonstrates the natural compatibility of tree-based planners with safety constraints and the robustness of the performance gain in out-of-training-domain scenarios.

\subsection{Dynamics Extension} 
As shown in Fig.~\ref{fig:NTE_general}, NTE can be applied to arbitrary game settings and dynamics. We evaluate the same Reach-Target-Avoid game with 3D Dubin's vehicle dynamics as a relevant model for fixed-wing aircraft applications, shown in Fig.~\ref{fig:dubins3d_state_space_ex}.
We consider the state, action, and dynamics: $\vs_t= [x_t,y_t,z_t,\psi_t,\gamma_t,\phi_t,v_t]^T$, $\va_t = [\dot{\gamma}_t,\dot{\phi}_t,\dot{v}_t]^T$
\begin{align}
    \vs_{t+1} &= \ff(\vs_t,\va_t) = 
   \vs_t + 
    \left[\begin{smallmatrix}
        v_t \cos(\gamma_t) \sin(\psi_t) \\ 
        v_t \cos(\gamma_t) \cos(\psi_t) \\ 
        - v_t \sin(\gamma_t) \\ 
        \frac{g}{v_t} \tan(\phi_t) \\ 
      \va_t
    \end{smallmatrix}\right] \cdelta_t
\end{align}
where $x,y,z$ are inertial position, $v$ is speed, $\psi$ is the heading angle, $\gamma$ is the flight path angle, and $\phi$ is the bank angle and $g$ is the gravitational acceleration.  The game is bounded to $\cpul = $\SI{5}{m} with a maximum linear acceleration of \SI{2.0}{m/s^2} and maximum angular rates of \SI{36}{deg/s}, and $g$ is set to \SI{0.98}{m/s\textsuperscript{2}} to scale to our game length scale.

\begin{figure}
    \centering 
    \includegraphics[width=0.4\textwidth]{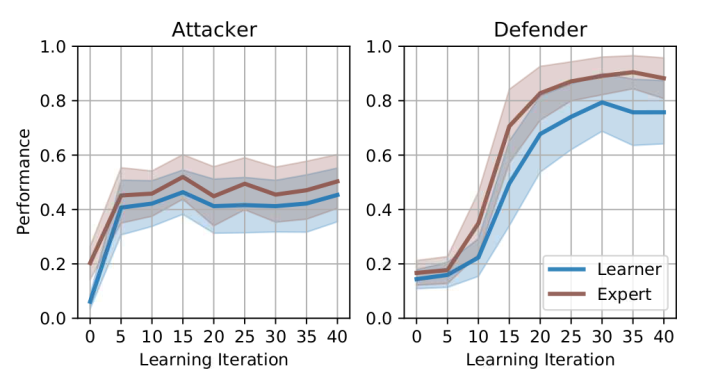}
    \caption{3D Dubin's vehicle game evaluation: the thick lines indicate the average performance and the shaded area is the variance over 100 games. 
    }
    \label{fig:exp3_3d_dubins}
\end{figure}

We initialize $2$ attacker, $2$ defender games for $100$ different initial conditions in a \SI{5}{m} region and test the policy variants, without an external baseline, for a total of \num{81 000} games. The performance results are shown in Fig.~\ref{fig:exp3_3d_dubins}, where we see the same trend that the learner and expert policies improve over learning iterations. In addition, the biased learner's performance quickly surpasses the unbiased expert ($k=0$). 


\subsection{Hardware Validation}
\label{sec:hardware}

To test our algorithm in practice, we fly in a motion capture space, where each robot (CrazyFlie 2.x,  see Fig.~\ref{fig:overview}) is equipped with a single marker, and we use the Crazyswarm~\cite{crazyswarm} for tracking and scripting. 
The centralized system simulates distributed operation by collecting the full state, computing local observations and local policies, and broadcasting only the output of each robot's learner policy. 
For a given double-integrator policy, we evaluate the learner to compute an action, forward-propagate double-integrator dynamics, and track the resulting position and velocity set-point using a nonlinear controller for full quadrotor dynamics.
Planning in a lower-dimensional double-integrator state and then tracking the full system is enabled by the timescale separation of position and attitude dynamics of quadrotors.

We evaluate the double-integrator learner for up to 3 attacker, 2 defender games in an aerial swarm flight demonstration. We show the results of the experiments in our supplemental video. We use the same parameters as in simulation in Sec.~\ref{sec:double_interator_evaluation}. Our learner evaluation takes an average of \SI{11}{ms} with a standard deviation of \SI{6}{ms}, with each robot policy process running in parallel on an Intel(R) Core(TM) i7-8665U. By comparison, the biased expert takes $329\pm$\SI{144}{ms} to execute and the unbiased expert takes $277\pm$\SI{260}{ms}. Our computational tests show that the learner has a significant ($\approx 25$ times) computational advantage over the baseline unbiased expert. 
Our physical demonstration shows that our learner is robust to the gap between simulation and real world and can run in real-time on off-the-shelf hardware. 

\section{Conclusion} 
\label{sec:conclusion}
We present a new approach for multi-robot planning in non-cooperative environments with an iterative search and learning method called Neural Tree Expansion. 
Our method bridges the gap between an AlphaZero-like method and real-world robotics applications by introducing a learner agent with decentralized evaluation, partial information, and limited computational resources. Our method outperforms the current state-of-the-art analytical baseline for the multi-robot double-integrator Reach-Target-Avoid game with dynamically sophisticated and coordinated strategies. 
We demonstrate our method's broad compatibility by further empirical evaluation of the Reach-Target-Avoid game for $3$D Dubin's vehicle dynamics and visualization of canonical decision-making problems. 
We validate the effectiveness through hardware experimentation and show that our policies run in real-time on off-the-shelf computational resources. 
In future work, we will combine planning under uncertainty algorithms with deep learning to handle scenarios with model and measurement uncertainty. 



\printbibliography

@article{paranjape2018robotic,
  title={Robotic herding of a flock of birds using an unmanned aerial vehicle},
  author={Paranjape, Aditya A and Chung, Soon-Jo and Kim, Kyunam and Shim, David Hyunchul},
  journal={IEEE Trans. Robot.},
  volume={34},
  number={4},
  pages={901--915},
  year={2018},
}

@inproceedings{Mansley_2011,
  author    = {Christopher R. Mansley and
              Ari Weinstein and
              Michael L. Littman},
  title     = {Sample-Based Planning for Continuous Action {Markov} Decision Processes, and Scheduling}, 
  booktitle = {{Int. Conf. on Autom. Planning and Scheduling}},
  year      = {2011},
}

@misc{Prajapat_2020,
%   author    = {Manish Prajapat and
%               Kamyar Azizzadenesheli and
%               Alexander Liniger and
%               Yisong Yue and
%               Anima Anandkumar},
%   title     = {Competitive Policy Optimization},
%   journal   = {CoRR},
%   volume    = {abs/2006.10611},
%   year      = {2020},
% %   archivePrefix = {arXiv},
% %   eprint = {0902.0885},
% }

@inproceedings{Sutton_1999,
  author    = {Richard S. Sutton and
               David A. McAllester and
               Satinder P. Singh and
               Yishay Mansour},
  title     = {Policy Gradient Methods for Reinforcement Learning with Function Approximation},
%   booktitle = {{NeurIPS}},
    booktitle = {{Neural Inf. Process. Syst.}},
%   booktitle = {{NIPS'99:} Proceedings of the 12\textsuperscript{th} International Conference on Neural Information Processing Systems},
  %pages     = {1057--1063},
  %publisher = {The {MIT} Press},
  year      = {1999}
}

@inproceedings{Kocsis_2006,
  author    = {Levente Kocsis and
               Csaba Szepesv{\'{a}}ri},
  title     = {Bandit Based {Monte-Carlo} Planning},
%   booktitle = {ECML},
  booktitle = {Eur. Conf. Mach. Learn.},
  %series    = {Lecture Notes in Computer Science},
  volume    = {4212},
  %pages     = {282--293},
  publisher = {Springer},
  year      = {2006}
}

@misc{Shah_2020,
%       title={Non-Asymptotic Analysis of {Monte Carlo} Tree Search}, 
%       author={Devavrat Shah and Qiaomin Xie and Zhi Xu},
%       year={2020},
%       eprint={1902.05213},
%       archivePrefix={arXiv},
% }

@Article{Shishika_2020,
  author  = {Daigo Shishika and James Paulos and Vijay Kumar},
  journal = {{IEEE} Robot. Autom. Lett.},
  title   = {Cooperative Team Strategies for Multi-Player Perimeter-Defense Games},
  year    = {2020},
  number  = {2},
  pages   = {2738--2745},
  volume  = {5},
  file    = {:papers/Shishika_2020.pdf:PDF},
%   url     = {https://ieeexplore.ieee.org/document/8988241},
}

@inproceedings{Chen_2020,
  author    = {Binghong Chen and
             others},
  title     = {Learning to Plan in High Dimensions via Neural Exploration-Exploitation
               Trees},
%   booktitle = {{ICLR}},
    booktitle = {{Int. Conf. on Learn. Repres.}},
  %publisher = {OpenReview.net},
  year      = {2020}
}

@inproceedings{Ichter_2018,
  author    = {Brian Ichter and
               James Harrison and
               Marco Pavone},
  title     = {Learning Sampling Distributions for Robot Motion Planning},
%   booktitle = {{IEEE ICRA}},
booktitle = {{Proc. IEEE Int. Conf. on Robot. and Automat.}},
%   booktitle = {{ICRA}},
  %pages     = {7087--7094},
%   publisher = {{IEEE}},
  year      = {2018}
}

@Book{Isaacs_1965,
    %   note = {Chapters 1-4 are revised versions of the author's reports  published as the Rand Corporation's Research memorandum,  RM-1391, RM-1399, RM-1411, RM-1486, all entitled in part  Differential games},
      author = {Isaacs, Rufus},
      publisher = {Wiley},
      year = 1965,
    %   url = {http://caltech.tind.io/record/497498},
      title = {Differential games; a mathematical theory with  applications to warfare and pursuit, control and  optimization},
    %  isbn      = {0486406822},
      %pages = {xxii, 384 p.},
}

@Article{garciaOptimalStrategiesClass2020,
  author       = {Garcia, Eloy and Casbeer, David W. and Pachter, Meir},
  title        = {Optimal Strategies for a Class of Multi-Player Reach-Avoid Differential Games in 3D Space},
%   issn         = {2377-3766},
  number       = {3},
  pages        = {4257--4264},
  volume       = {5},
%   abstract     = {A multi-player reach-avoid differential game with autonomous aerial robots in the three dimensional space is studied. Two pursuers form a team to guard a target against an evader with the same speed as the pursuers. This letter provides the complete solution of this differential game that resides within a high-dimensional state space. The Barrier surface is characterized and the saddle-point strategies are synthesized and verified. Degeneration of the two-pursuer one-evader game into the one-on-one case is addressed and the corresponding strategies are obtained. Finally, several examples illustrate the robustness properties and the guarantees provided by the saddle-point strategies obtained in this letter.},
  date         = {2020},
%   doi          = {10.1109/LRA.2020.2994023},
%   eventtitle   = {{{IEEE Robotics}} and {{Automation Letters}}},
%   file         = {:papers/garciaOptimalStrategiesClass2020.pdf:PDF;:/home/whoenig/Zotero/storage/I98AGB52/Garcia et al. - 2020 - Optimal Strategies for a Class of Multi-Player Rea.pdf:;:/home/whoenig/Zotero/storage/NK8VT54K/9091321.html:},
%   journaltitle = {IEEE Robotics and Automation Letters},
  journal   = {{IEEE} Robot. Autom. Lett.},
%   keywords     = {3D space,aerial systems: applications,autonomous aerial robots,cooperating robots,differential games,Differential games,Game theory,Games,high-dimensional state space,mobile robots,multiplayer reach-avoid differential game,Optimal control,optimal strategies,optimisation,optimization and optimal control,Robots,Robustness,saddle-point strategies,Three-dimensional displays,two-pursuer one-evader game,Unmanned aerial vehicles},
}

@article {Silver1140,
	author = {Silver, David and others},
	title = {A general reinforcement learning algorithm that masters chess, shogi, and Go through self-play},
	volume = {362},
	number = {6419},
	pages = {1140--1144},
	year = {2018},
% 	doi = {10.1126/science.aar6404},
	publisher = {American Association for the Advancement of Science},
	journal = {Science}
}

@InProceedings{coonControlStrategiesMultiplayer2017,
  author     = {Coon, Mitchell and Panagou, Dimitra},
  booktitle  = {{{IEEE}} 56th {{Annual Conf.}} on {{Decis.}} and {{Control}}},
%   booktitle  = {IEEE {CDC}},
  year       = 2017,
  title      = {Control Strategies for Multiplayer Target-Attacker-Defender Differential Games with Double Integrator Dynamics},
  %pages      = {1496--1502},
  abstract   = {This paper presents a method for deriving optimal controls and assigning attacker-defender pairs in a target-attacker-defender differential game between an arbitrary numbers of attackers and defenders, all of which are modeled using double integrator dynamics. It is assumed that each player has perfect information about the states and controls of the players within a certain range of themselves, but they are unaware of any players outside of this range. Isochrones are created based on the time-optimal trajectories needed for the players to reach any point in the shortest possible time. The intersections of the players' isochrones are used to determine whether a defender can intercept an attacker before the attacker reaches the target. Sufficient conditions on the detection range of the defenders and the guaranteed capture despite perturbations of the attackers off the nominal trajectories are derived. Then, in simulations with multiple players, attacker-defender pairs are assigned so that the maximum number of attackers are intercepted in the shortest possible time.},
  %date       = {2017-12},
%   doi        = {10.1109/CDC.2017.8263864},
  eventtitle = {2017 {{IEEE}} 56th {{Annual Conference}} on {{Decision}} and {{Control}} ({{CDC}})},
  file       = {:papers/coonControlStrategiesMultiplayer2017.pdf:PDF;:/home/whoenig/Zotero/storage/VM67JGSK/Coon and Panagou - 2017 - Control strategies for multiplayer target-attacker.pdf:;:/home/whoenig/Zotero/storage/JQJVDB3J/8263864.html:},
  keywords   = {Acceleration,Aircraft,Atmospheric modeling,attacker-defender pairs,differential games,double integrator dynamics,Games,multiplayer target-attacker-defender differential games,optimal control,Optimal control,optimal controls,time-optimal trajectories,Trajectory,Vehicle dynamics},
}

@inproceedings{Ross_2011,
  author    = {St{\'{e}}phane Ross and
               Geoffrey J. Gordon and
               Drew Bagnell},
  title     = {A Reduction of Imitation Learning and Structured Prediction to No-Regret Online Learning},
%   booktitle = {{AISTATS}},
booktitle = {{Proc.  Int.  Conf.  Artif.  Intell. and  Statist.}},
%   series    = {{JMLR} Proceedings},
%  volume    = {15},
%   pages     = {627--635},
%   publisher = {JMLR.org},
  year      = {2011}
}

@Article{Yan_2019,
%   author      = {Rui Yan and Xiaoming Duan and Zongying Shi and Yisheng Zhong and Francesco Bullo},
%   title       = {Matching-Based Capture Strategies for {3D} Heterogeneous Multiplayer Reach-Avoid Differential Games},
%   %date        = {2019-09-26},
%   year        = {2019},
% %   eprint      = {1909.11881},
%   %eprintclass = {math.OC},
%   eprinttype  = {arXiv},
% %   file        = {:papers/Yan_2019.pdf:PDF},
%   %keywords    = {math.OC},
% %   url         = {https://arxiv.org/pdf/1909.11881.pdf},
% }

@incollection{pyTorch,
title = {PyTorch: An Imperative Style, High-Performance Deep Learning Library},
author = {Adam Paszke and others},
%booktitle = {Advances in Neural Information Processing Systems 32},
% booktitle = {{NeurIPS}},
    booktitle = {{Neural Inf. Process. Syst.}},
editor_ = {H. Wallach and H. Larochelle and A. Beygelzimer and F. d\textquotesingle Alch\'{e}-Buc and E. Fox and R. Garnett},
% pages = {8024--8035},
year = {2019},
publisher_ = {Curran Associates, Inc.},
% url = {http://papers.neurips.cc/paper/9015-pytorch-an-imperative-style-high-performance-deep-learning-library.pdf}
}

@article{Browne_2012,
  author    = {Cameron Browne and others},
  title     = {A Survey of {Monte Carlo} Tree Search Methods},
  journal   = {{IEEE} Trans. Comput. Intell. {AI} Games},
  volume    = {4},
  number    = {1},
  pages     = {1--43},
  year      = {2012}
}

@inproceedings{Zaheer_2017,
  author    = {Manzil Zaheer and
              others},
  title     = {Deep Sets},
%   booktitle = {{NeurIPS}},
    booktitle = {{Neural Inf. Process. Syst.}},
%   pages     = {3391--3401},
  year      = {2017}
}

@inproceedings{Moerland_2018,
  author    = {Thomas M. Moerland and
               Joost Broekens and
               Aske Plaat and
               Catholijn M. Jonker},
  title     = {{A0C:} {Alpha} Zero in Continuous Action Space},
  booktitle = {Eur. Workshop on Reinforcement Learn. 14},
  year      = 2018, 
%   European Workshop on Reinforcement Learning 14 (2018)
%   archivePrefix = {arXiv},
%       eprint = {1805.09613},
  %journal   = {CoRR},
  %volume    = {abs/1805.09613},
  %year      = {2018}
}

@inproceedings{crazyswarm,
  author    = {James A. Preiss and
               Wolfgang  H\"onig and
               Gaurav S. Sukhatme and
               Nora Ayanian},
  title     = {Crazyswarm: {A} large nano-quadcopter swarm},
  booktitle = {Proc. IEEE Int. Conf. Robot. Autom.},
  %pages     = {3299--3304},
  publisher_ = {{IEEE}},
  year      = {2017},
%   url       = {https://doi.org/10.1109/ICRA.2017.7989376},
%   doi_       = {10.1109/ICRA.2017.7989376},
}

@article{Riviere_2020,
  author    = {Benjamin Rivi{\`{e}}re and
               Wolfgang H{\"{o}}nig and
               Yisong Yue and
               Soon{-}Jo Chung},
  title     = {{GLAS:} {G}lobal-to-Local Safe Autonomy Synthesis for Multi-Robot Motion
               Planning With End-to-End Learning},
  journal   = {{IEEE} Robot. Autom. Lett.},
  volume    = {5},
  number    = {3},
  pages     = {4249--4256},
  year      = {2020}
}

@inproceedings{Mao_2020,
  author    = {W. Mao and
               K. Zhang and
               Q. Xie and
               T. Basar},
  title     = {{POLY-HOOT:} {Monte-Carlo} Planning in Continuous Space {MDP}s with Non-Asymptotic
               Analysis},
%   booktitle = {{NeurIPS}},
    booktitle = {{Neural Inf. Process. Syst.}},
  year      = {2020}
}

@inproceedings{Sunberg_2018,
  author    = {Z. N. Sunberg and
               M. J. Kochenderfer},
  title     = {Online Algorithms for {POMDPs} with Continuous State, Action, and Observation
               Spaces},
%   booktitle = {{ICAPS}},
  booktitle = {{Int. Conf. on Autom. Planning and Scheduling}},
  %pages     = {259--263},
  %publisher = {{AAAI} Press},
  year      = {2018}
}

@inproceedings{Florensa_2017,
  author    = {Carlos Florensa and
               David Held and
               Markus Wulfmeier and
               Michael Zhang and
               Pieter Abbeel},
  title     = {Reverse Curriculum Generation for Reinforcement Learning},
%   booktitle = {CoRL},
booktitle = {Conf. on Robot Learn.},
%   series    = {Proceedings of Machine Learning Research},
  volume    = {78},
  %pages     = {482--495},
%   publisher = {{PMLR}},
  year      = {2017}
}

@inproceedings{Auger_2013,
  author    = {David Auger and
               Adrien Cou{\"{e}}toux and
               Olivier Teytaud},
  title     = {Continuous Upper Confidence Trees with Polynomial Exploration -- {Consistency}},
  booktitle = {{Eur. Conf. Mach. Learn.}},
  %series    = {Lecture Notes in Computer Science},
  volume    = {8188},
%   pages     = {194--209},
  publisher = {Springer},
  year      = {2013}
}

@ARTICLE{Hu_2020,
  author={Hu, Junyan and Turgut, Ali Emre and Krajník, Tomáš and Lennox, Barry and Arvin, Farshad},
  journal={IEEE Trans. on Cogn. and Develop. Syst.}, 
  title={Occlusion-Based Coordination Protocol Design for Autonomous Robotic Shepherding Tasks}, 
  year={2020},
  volume={},
  number={},
  pages={1-1},
%   doi={10.1109/TCDS.2020.3018549}
}

@article{Nardi_2018,
  author    = {Simone Nardi and
               Federico Mazzitelli and
               Lucia Pallottino},
  title     = {A Game Theoretic Robotic Team Coordination Protocol For Intruder Herding},
  journal   = {{IEEE} Robot. Autom. Lett.},
  volume    = {3},
  number    = {4},
  pages     = {4124--4131},
  year      = {2018}
}

@book{Kochenderfer_2015,
author = {Kochenderfer, Mykel J. and others},
title = {Decision Making Under Uncertainty: Theory and Application},
year = {2015},
publisher = {The MIT Press},
edition = {1\textsuperscript{st} Ed.}
}

\end{document}